\documentclass[11pt]{article}
\usepackage[margin=1in]{geometry}
\usepackage{amsmath,amssymb}
\usepackage[table]{xcolor}
\usepackage{booktabs}
\usepackage{siunitx}
\usepackage{hyperref}

\usepackage{amsthm}
\newtheorem*{remark}{Remark}

\definecolor{tableblue}{HTML}{5A86A3}
\newcommand{\sh}[2]{\cellcolor{tableblue!#1}#2}

\title{Ornstein–Uhlenbeck Is Hard to Beat, Yet Superlinear Drift Ships Lower Transport Costs}
\author{}
\date{}

\author{
  Attila Lovas \qquad L\'or\'ant Nagy\\[0.4em]
  \small HUN-REN Alfr\'ed R\'enyi Institute of Mathematics\\
  \small Budapest, Hungary
}

\begin{document}
\maketitle

\begin{abstract}
Bre\v{s}ar and Mijatovi\'c \cite{bresar2025} show that Ornstein--Uhlenbeck diffusion is hard to beat in forward convergence under assumptions that exclude superlinear drift. We instead test superlinear Langevin diffusions for score-based image generation, computing their conditional scores numerically from a Fokker--Planck equation. In our experiments, the superlinear models beat the Ornstein--Uhlenbeck baseline on empirical Wasserstein distance across nearly the entire tested grid and show less variation across diffusion horizons. The ``hard to beat'' verdict of \cite{bresar2025} thus fails to be universal.
\end{abstract}

\section{Introduction}

Let $\mu_\star$ be a data distribution on $\mathbb{R}^d$. Diffusion-based generative modelling introduces a forward stochastic process that progressively destroys the structure of $\mu_\star$, and then constructs a reverse process that transports a simple terminal law back toward the data distribution. This principle appears in the diffusion model of Sohl-Dickstein et al.\ \cite{sohldickstein2015}, the denoising diffusion model of Ho et al.\ \cite{ho2020}, and the continuous-time score-based formulation of Song et al.\ \cite{song2021}. The reverse-time diffusion formula itself is classical \cite{anderson1982,haussmann1986}, and score matching provide the learning principle.

Common continuous-time constructions use affine Gaussian corruptions whose conditional transition laws and conditional scores are available analytically. The present experiment replaces the linear mean-reverting drift by a coordinatewise superlinear Langevin drift. The transition score is then computed numerically from the associated Fokker--Planck equation and used as the regression target for the score network.

The construction does not require an analytical expression for either the corrupted-data marginal $p_t$ or the conditional transition density $p_{t|0}$. When the forward diffusion has a sufficiently regular transition density, its conditional score can instead be approximated by solving the associated Fokker--Planck equation. This permits forward processes beyond those with closed-form Gaussian transitions. In the present experiment we restrict attention to coordinatewise Langevin diffusions, for which the conditional score can be computed from a one-dimensional equation and reused across coordinates.

\section{Forward diffusion}
\label{sec:forward}

Fix $\alpha\geq1$ and $c_\alpha>0$. Define the scalar potential
\[
U_\alpha(y)
=
\frac{c_\alpha}{\alpha+1}|y|^{\alpha+1},
\]
and the potential on $\mathbb R^d$ by
\[
\mathcal U_\alpha(x)
=
\sum_{i=1}^d U_\alpha(x^i).
\]
The forward process is the gradient Langevin diffusion
\begin{equation}
dX_t
=
B_\alpha(X_t)\,dt
+dW_t,
\qquad
X_0\sim\mu_\star,
\quad
t\in[0,T].
\label{eq:forward}
\end{equation}
where
\[
B_\alpha(x)
=
-\nabla\mathcal U_\alpha(x).
\]
Coordinatewise we use the notation,
\[
B_\alpha(x)
=
\bigl(
b_\alpha(x^1),\ldots,b_\alpha(x^d)
\bigr),
\]
where by definition
\[
b_\alpha(y)
=
-U_\alpha'(y)
=
-c_\alpha |y|^\alpha\operatorname{sgn}(y).
\]

\noindent Define
\[
Z_\alpha
=
\int_{\mathbb R^d}
\exp\!\left[-2\mathcal U_\alpha(x)\right]\,dx,
\qquad
V_\alpha(x)
=
2\mathcal U_\alpha(x)+\log Z_\alpha.
\]
The invariant probability measure of \eqref{eq:forward} is
\begin{equation}
\Pi_\alpha(dx)
=
e^{-V_\alpha(x)}\,dx
=
Z_\alpha^{-1}
\exp\!\left[-2\mathcal U_\alpha(x)\right]\,dx.
\label{eq:invariant}
\end{equation}

\noindent The coordinates evolve independently conditional on the initial image. For $y_0\in\mathbb R$, let $q^{y_0}(t,y)$ denote the transition density of the scalar diffusion
\[
dY_t
=
b_\alpha(Y_t)\,dt
+dB_t,
\qquad
Y_0=y_0.
\]
Then the conditional transition density of \eqref{eq:forward} factorizes as
\begin{equation}
p_{t|0}(x\mid x_0)
=
\prod_{i=1}^d q^{x_0^i}(t,x^i).
\label{eq:factorization}
\end{equation}

\section{Time reversal for the superlinear family}

We now justify the reverse-time dynamics of the continuous-time process
\eqref{eq:forward} using the findings in \cite{cattiaux2023}. Assume that
\begin{equation}
H(\mu_\star\mid\Pi_\alpha)<\infty.
\label{eq:entropyassumption}
\end{equation}

\noindent Let $Y$ follow the same Langevin dynamics as $X$, with an  invariant initial law:
\[
dY_t
=
B_\alpha(Y_t)\,dt
+dW_t,
\qquad
Y_0\sim\Pi_\alpha.
\]
Denote the path laws of $X$ and $Y$ by $P$ and $R$, respectively. In the notation of \cite{cattiaux2023}, the diffusion matrix in the present setup is
\[
a=I_d.
\]
Since
$V_\alpha(x)
=
2\mathcal U_\alpha(x)+\log Z_\alpha,
$
we have
$
\nabla V_\alpha(x)
=
2\nabla\mathcal U_\alpha(x),
$
and therefore
$
B_\alpha(x)
=
-\frac12a\nabla V_\alpha(x)
$. Thus $R$ is the reversible diffusion associated with the invariant measure
\[
\Pi_\alpha(dx)=e^{-V_\alpha(x)}\,dx.
\]

\noindent Moreover, $V_\alpha\in C^1(\mathbb R^d)$, $a=I_d$ is constant, and the growth condition of \cite{cattiaux2023} is satisfied. Indeed,
\[
\begin{aligned}
x\cdot B_\alpha(x)+\operatorname{tr}(a)
&=
-c_\alpha
\sum_{i=1}^d
x^i|x^i|^\alpha\operatorname{sgn}(x^i)
+d\\
&=
-c_\alpha
\sum_{i=1}^d
|x^i|^{\alpha+1}
+d\\
&\le d.
\end{aligned}
\]

The processes $X$ and $Y$ have the same conditional path law given their initial point and differ only in their initial distributions. Consequently,
\[
H(P\mid R)
=
H(\mu_\star\mid\Pi_\alpha)
<
\infty.
\]
Thus the reference diffusion satisfies the required nondegeneracy, reversibility, and growth conditions moreover, $P$ is Markov and, under assumption \eqref{eq:entropyassumption}, has finite relative entropy with respect to $R$. The time-reversal theorem of \cite{cattiaux2023} therefore applies, and hence the stationary process $Y$ is reversible.

\begin{remark}
The finite-entropy assumption \eqref{eq:entropyassumption} can be enforced by an arbitrarily small Gaussian regularization of the data distribution. Indeed, if $\mu_\star$ is supported on a bounded subset of $\mathbb R^d$, as is the case for normalized image data, then for any $\varepsilon>0$ the regularized law
\[
\mu_\star^\varepsilon
=
\mu_\star * \mathcal N(0,\varepsilon^2 I_d)
\]
has a smooth strictly positive density and satisfies
\[
H(\mu_\star^\varepsilon\mid\Pi_\alpha)<\infty.
\]
Thus the assumption \eqref{eq:entropyassumption} may be ensured by adding an arbitrarily small amount of independent Gaussian noise to the data.
\end{remark}

\section{Reverse diffusion and the score}
\label{sec:reverse}

Let $p_t$ denote the density of $X_t$, and define reverse time by
\[
\tau=T-t,
\qquad
\bar X_\tau=X_{T-\tau}.
\]
Under the finite-entropy assumption \eqref{eq:entropyassumption}, the argument in the preceding section gives
\begin{equation}
d\bar X_\tau
=
\left[
-B_\alpha(\bar X_\tau)
+
\nabla\log p_{T-\tau}(\bar X_\tau)
\right]d\tau
+
d\bar W_\tau,
\qquad
\bar X_0\sim\operatorname{Law}(X_T).
\label{eq:reverse}
\end{equation}

As usual, the unknown quantity in this equation is the marginal score
\[
\nabla\log p_t(x).
\]
Although the conditional forward dynamics decouple coordinatewise, the marginal law at time $t$ generally does not, because the initial distribution $\mu_\star$ contains dependencies between image coordinates. Indeed,
\[
p_t(x)
=
\int_{\mathbb R^d}
p_{t|0}(x\mid x_0)\,\mu_\star(dx_0).
\]
The marginal score at $x$ depends on the posterior distribution of the original image $X_0$ given $X_t=x$. A neural network $s_\theta(x,t)$ is therefore trained to approximate the full $d$-dimensional marginal score. The regression target is the conditional score. The denoising score-matching objective is
\begin{equation}
\mathcal L(\theta)
=
\mathbb E\left[
\left\|
s_\theta(X_t,t)
-
\nabla_x\log p_{t|0}(X_t\mid X_0)
\right\|_2^2
\right].
\label{eq:dsm}
\end{equation}
Under the usual regularity assumptions,
\begin{equation}
\mathbb E\left[
\nabla_x\log p_{t|0}(X_t\mid X_0)
\mid X_t=x
\right]
=
\nabla_x\log p_t(x),
\label{eq:denoiseidentity}
\end{equation}
so the population minimizer of \eqref{eq:dsm} is the marginal score required in \eqref{eq:reverse} \cite{vincent2011,song2021}.

\section{Score construction and numerical implementation}
\label{sec:implementation}

The time-reversal formula specifies the reverse drift, but its marginal
score $\nabla\log p_t$ is not available in closed form. To train the
network, we instead construct the conditional score in
\eqref{eq:dsm}. For fixed $y_0$, the scalar transition density
$q^{y_0}(t,y)$ defined in Section~\ref{sec:forward} satisfies the
Fokker--Planck equation
\begin{equation}
\partial_t q
=-\partial_y\bigl(b_\alpha(y)q\bigr)
+\frac{1}{2}\partial_y^2q,
\qquad q(0,\cdot)=\delta_{y_0}.
\label{eq:fp}
\end{equation}
Consequently, the conditional score has components
\begin{equation}
s(y_0,t,y)=\partial_y\log q^{y_0}(t,y),
\qquad
\bigl[\nabla_x\log p_{t|0}(x\mid x_0)\bigr]^i
=s(x_0^i,t,x^i).
\label{eq:scalarscore}
\end{equation}
For the nonlinear choices of $\alpha$, we approximate $q^{y_0}$
numerically rather than use a closed-form transition density
\cite{risken1989,bogachev2015}.

We solve \eqref{eq:fp} on a finite grid for a range of starting values
$y_0$. A narrow Gaussian approximates the initial point mass. We then take finite
differences of the log-density and store the resulting conditional
scores in a table indexed by $(y_0,t,y)$. During training, scores are
obtained from this table by trilinear interpolation.

For the experiments, the data consist of $28\times28$ grayscale images
normalized to $[-1,1]^d$, where $d=784$. We set
$c_\alpha=0.5$, $c_0=0$, and test
$\alpha\in\{1,2,3\}$, $T\in\{3,\ldots,10\}$, and
$N\in\{100,150,200\}$. The score table uses $200$ starting-value
points in $[-1.5,1.5]$, $700$ time points in $[0,T]$, and $200$ state
points in $[-5.5,5.5]$. The initial Gaussian has standard deviation
$0.03$, and densities are floored at $10^{-12}$ before taking
logarithms. For $\alpha=1$, the forward process is
$dX_t=-\tfrac12X_t\,dt+dW_t$ and its invariant law is
$\mathcal N(0,I_d)$.

Training samples of the forward process are generated by
Euler--Maruyama with step size $h=T/N$:
\begin{equation}
\widehat X_{k+1}
=\widehat X_k+hB_\alpha(\widehat X_k)
+\sqrt h\,\xi_k,
\qquad \xi_k\sim\mathcal N(0,I_d).
\label{eq:em}
\end{equation}
For each image, one nonzero time is selected uniformly from its
$N$-step trajectory. A U-Net is trained to predict the tabulated
conditional score at that time. It has $64$ base channels, channel
multipliers $(1,2)$, two residual blocks per level, dropout $0.1$,
and a time embedding of dimension $128$. We use AdamW for $15$
epochs with batch size $64$, learning rate $10^{-5}$, and zero weight
decay. 

Reverse sampling also uses Euler--Maruyama with $N$ steps. Its
starting law is approximated by the invariant law $\Pi_\alpha$;
samples from that law are approximated by evolving standard normal
samples under the forward Langevin equation to time
$T_{\mathrm{eq}}=30$. This equilibration uses $N$ steps of size
$30/N$. For superlinear drifts, explicit Euler can produce rare,
very large excursions \cite{hutzenthaler2011}; numerical explosions
are monitored, and generated images with more than ten percent
non-finite coordinates are discarded at evaluation time.

\bigskip
\noindent
The code used for the numerical experiments is available at
\url{https://github.com/lorant-nagy/genai}.

\section{Evaluation and results}
\label{sec:res}

We evaluate generated images using the empirical Wasserstein--1 distance with Euclidean ground cost. Real and generated $28\times28$ images are mapped to $[0,1]^d$ and flattened, a fixed set of $3000$ real images is used. The distance is computed by solving the discrete optimal-transport problem \cite{flamary2021}.

Across the tested grid, a superlinear choice ($\alpha=2$ or $\alpha=3$) gives a lower value than the linear baseline in $23$ of the $24$ fixed $(T,N)$ comparisons; the exception is $N=100$, $T=8$. The best value in each table is also attained by a superlinear model, and the row summaries show less variation across $T$. We conjecture that stronger mean reversion contributes to both the improved Wasserstein--1 scores and their reduced sensitivity to the choice of $T$.

\bigskip

Our findings qualify the ``hard to beat'' characterization of the Ornstein--Uhlenbeck process in \cite{bresar2025}: the results therein concern forward convergence to stationarity under at-most-linear drift, while our superlinear models achieve lower empirical generation error in the experiments reported here.

\clearpage
\begin{table}[p]
\centering
\small

\begin{tabular}{c *{8}{S[table-format=2.4]}}
\toprule
& \multicolumn{8}{c}{$T$} \\
\cmidrule(l){2-9}
$\alpha$ & {$3$} & {$4$} & {$5$} & {$6$} & {$7$} & {$8$} & {$9$} & {$10$} \\
\midrule
$3$ & \sh{42}{7.4895} & \sh{45}{7.1817} & \sh{45}{7.2128} & \sh{43}{7.3525} & \sh{42}{7.5041} & \sh{41}{7.6127} & \sh{40}{7.6784} & \sh{40}{7.7590} \\
$2$ & \sh{37}{8.0436} & \sh{45}{\bfseries 7.1735} & \sh{45}{7.1995} & \sh{44}{7.2805} & \sh{43}{7.4185} & \sh{42}{7.5190} & \sh{40}{7.7112} & \sh{41}{7.6554} \\
$1$ & \sh{12}{10.7980} & \sh{37}{8.1018} & \sh{23}{9.5526} & \sh{43}{7.4124} & \sh{43}{7.4386} & \sh{43}{7.4421} & \sh{40}{7.7372} & \sh{40}{7.7217} \\
\bottomrule
\end{tabular}
\caption{$\mathcal W_1$ for $N=100$. Lower is better; darker shading indicates a smaller value.}
\label{tab:n100}

\vspace{0.6em}

\begin{tabular}{c *{8}{S[table-format=2.4]}}
\toprule
& \multicolumn{8}{c}{$T$} \\
\cmidrule(l){2-9}
$\alpha$ & {$3$} & {$4$} & {$5$} & {$6$} & {$7$} & {$8$} & {$9$} & {$10$} \\
\midrule
$3$ & \sh{41}{7.6303} & \sh{44}{7.2553} & \sh{45}{\bfseries 7.2215} & \sh{43}{7.4292} & \sh{42}{7.4934} & \sh{42}{7.4947} & \sh{41}{7.5667} & \sh{38}{7.9462} \\
$2$ & \sh{35}{8.3078} & \sh{44}{7.2908} & \sh{43}{7.4071} & \sh{43}{7.4038} & \sh{42}{7.4565} & \sh{42}{7.5470} & \sh{42}{7.5285} & \sh{36}{8.1154} \\
$1$ & \sh{9}{11.1001} & \sh{8}{11.1918} & \sh{26}{9.2831} & \sh{41}{7.5673} & \sh{12}{10.7467} & \sh{38}{7.9927} & \sh{34}{8.3327} & \sh{20}{9.8957} \\
\bottomrule
\end{tabular}
\caption{$\mathcal W_1$ for $N=150$. Lower is better; darker shading indicates a smaller value.}
\label{tab:n150}

\vspace{0.6em}

\begin{tabular}{c *{8}{S[table-format=2.4]}}
\toprule
& \multicolumn{8}{c}{$T$} \\
\cmidrule(l){2-9}
$\alpha$ & {$3$} & {$4$} & {$5$} & {$6$} & {$7$} & {$8$} & {$9$} & {$10$} \\
\midrule
$3$ & \sh{35}{8.3180} & \sh{45}{\bfseries 7.2164} & \sh{44}{7.3147} & \sh{43}{7.3457} & \sh{43}{7.3593} & \sh{41}{7.5742} & \sh{39}{7.8070} & \sh{41}{7.6255} \\
$2$ & \sh{24}{9.4270} & \sh{42}{7.4931} & \sh{42}{7.5542} & \sh{42}{7.5268} & \sh{43}{7.3764} & \sh{42}{7.5384} & \sh{36}{8.1468} & \sh{36}{8.1259} \\
$1$ & \sh{9}{11.1501} & \sh{32}{8.5657} & \sh{25}{9.3656} & \sh{5}{11.5609} & \sh{13}{10.7299} & \sh{18}{10.1254} & \sh{22}{9.6737} & \sh{35}{8.2259} \\
\bottomrule
\end{tabular}
\caption{$\mathcal W_1$ for $N=200$. Lower is better; darker shading indicates a smaller value.}
\label{tab:n200}

\vspace{0.6em}

\begin{tabular}{c c c c}
\toprule
$N$ & $\alpha=1$ & $\alpha=2$ & $\alpha=3$ \\
\midrule
$100$ & $7.92\pm0.76$ & $7.42\pm0.22$ & $7.47\pm0.23$ \\
$150$ & $9.29\pm1.39$ & $7.54\pm0.27$ & $7.49\pm0.24$ \\
$200$ & $9.75\pm1.17$ & $7.68\pm0.32$ & $7.46\pm0.21$ \\
\bottomrule
\end{tabular}
\caption{Mean and sample standard deviation over $T\in\{4,\ldots,10\}$.}
\label{tab:rows}

\end{table}

\end{document}